%% file: ef-kdd.tex
\documentclass[sigconf]{acmart}
\usepackage{booktabs} 
\usepackage{stmaryrd}
\usepackage{amssymb}
\usepackage{graphicx}
\usepackage{caption}
\usepackage{subcaption}
\usepackage{balance}

\DeclareMathAlphabet\mathbfcal{OMS}{cmsy}{b}{n}
\setcopyright{rightsretained}

\acmDOI{10.475/123_4}

\acmISBN{123-4567-24-567/08/06}

\acmConference[KDD'17]{ACM KDD 2017}{August 2017}{Nova Scotia, Canada} 
\acmYear{2017}
\copyrightyear{2017}

\acmPrice{15.00}

\begin{document}
\title{Non-negative Tensor Factorization for Human Behavioral Pattern Mining in Online Games}

\author{Anna Sapienza}
\affiliation{University of Southern California\\Information Sciences Institute}
\email{annas@isi.edu}

\author{Alessandro Bessi}
\affiliation{University of Southern California\\Information Sciences Institute}
\email{bessi@isi.edu}

\author{Emilio Ferrara}
\affiliation{University of Southern California\\Information Sciences Institute}
\email{emiliofe@usc.edu}

\begin{abstract}
Multiplayer online battle arena has become a popular game genre. It also received increasing attention from our research community because they provide a wealth of information about human interactions and behaviors. A major problem is extracting meaningful patterns of activity from this type of data, in a way that is also easy to interpret. Here, we propose to exploit tensor decomposition techniques, and in particular Non-negative Tensor Factorization, to discover hidden correlated behavioral patterns of play in a popular game: League of Legends. We first collect the entire gaming history of a group of about one thousand players, totaling roughly $100K$ matches.
By applying our methodological framework, we then separate players into groups that exhibit similar features and playing strategies, as well as similar temporal trajectories, i.e., behavioral progressions over the course of their gaming history: this will allow us to investigate how players learn and improve their skills.
\end{abstract}

%
%




\maketitle

\section{introduction}

Multiplayer Online Battle Arena (MOBA) is a genre of strategy online games that has drawn growing attention and has become extremely popular. MOBA consist of match-based games, where players, divided in two opposing teams, compete against each other. Each player during the match controls a single character (a.k.a., hero), having a specific role and abilities.

The main goal in each match is to destroy the enemy team's base, while enhancing the player level, increasing the abilities of the controlled character, and cooperating with one's own teammates.

This genre of games, including \textit{Heroes of the Storm}, \textit{Dota 2}, and \textit{League of Legends}, has attracted researchers from different fields, especially because they provide a unique way to study the influence of role-playing in competitive games~\cite{ferrari2013generative,foo2004defining}, the impact of cooperation~\cite{kou2014playing,brown2004cscw} versus individual player attitudes~\cite{nuangjumnong2014effects}, social behaviors~\cite{kou2013regulating,ducheneaut2004social,shores2014identification}, user commitment~\cite{dabbish2012communication}, etc.

The analysis of MOBA games also allows for the discovery of useful information to study the social dynamics of player communities. For example, by extracting players' social activities and relations, researchers addressed issues such as gender gap~\cite{ratan2015stand,ratan2012league} and improved the user experience~\cite{veron2014matchmaking}. 

One advantage of analyzing players' records in online games is the possibility of monitoring how their behaviors evolve over time. The temporal dimension exhibited by such data enables the study of the evolution of player performance, specifically how players learn, adapt, and modify their playing strategies over time. 

We propose to study both the temporal and social dynamics of players in MOBA games at once. 
Here, we will focus on the analysis of League of Legends (LoL), a popular MOBA game. 
Our goal will be that of identifying different groups of players with common strategies, such as collaborative versus individualist players, and understanding how these groups of players behave in time, i.e., how their strategies evolve over time. 

To this aim, we take advantage of Non-negative Tensor Factorization (NTF)~\cite{cichocki2009nonnegative,lim2009nonnegative} techniques, which derive from multi-linear algebra. Non-negative tensor factorization allows to identify correlation in the data at different levels~\cite{kolda2009tensor}: on the one hand, the application of the NTF helps in the identification of hidden topological structures in the data, like groups or communities, which are easy to interpret as they reflect individuals' social dynamics~\cite{gauvin2014detecting}. On the other hand, these topological structures share correlated activity patterns~\cite{sapienza2015detecting,panisson2014mining}.  Our purpose is therefore to leverage NTF to detect groups of players characterized by similar features and strategies as well as their temporal trajectories, i.e., their evolution. 


\subsection*{Contributions}

Through the lens of NTF, we study the gaming history of about one thousand League of Legends players accounting for a total of roughly $100K$ matches. Our analysis will: 

\begin{itemize}
\item Highlight the existence of an underlying structure in the data, that allows us to divide players in groups characterized by similar features and having correlated temporal behaviors;
\item Provide an interpretation of the components extracted by the NTF;
\item Validate the interpretation of the NTF results by analyzing the uncovered groups and their evolution over time;
\item Discover, by analyzing the temporal components, that players' playing strategies are consistent over time;
\item Provide and validate an explanation for players behavioral stability, namely that the design of the game strongly impacts team formation in each match, thus manipulating the team's probability of victory.
\end{itemize}


\begin{table}[h]
\centering
\caption{Notation used throughout the text.}
\label{notation}
\begin{tabular}{c|c}
Notation & Definition \\ \hline
$X$ & constant \\ \hline
$x$ & scalar \\
\hline
$\mathbf{x}$ & vector \\
\hline
$\mathbf{X}$ & matrix \\
\hline
$x_{ij}$ & matrix entry \\
\hline
$\mathbfcal{X}$ & tensor \\
\hline
$x_{ijk}$ & entry of a three-dimensional tensor \\
\hline
$\circ$ & outer product
\end{tabular}
\end{table}

\section{methodology}
\label{met}
The extraction of meaningful patterns of behavior in online games can be carried out by taking full advantage of tensor decomposition techniques. These provide a way to disentangle the temporal and topological characteristics of the studied system. This is achieved by performing a dimensionality reduction on the original system. 

Let us consider a dataset composed by users (i.e., players), whose characteristics (i.e., features) evolve over time.
To perform the decomposition of such dataset, we need to represent it as a tensor (i.e., a multi-dimensional array). This can be done by assigning to each dimension of the tensor the different dimensions of the data, namely users, features, and time. In this context, the time axis corresponds to LoL matches.
Following the notation reported in Table~\ref{notation},  we thus define a three-dimensional array, denoted as $\mathbfcal{X}\in\mathbb{R}^{I\times J\times K}$, where $I$ is the number of users, $J$ is the number of features, and $K$ is the number of time steps (i.e., played matches). In this formulation, the entry $x_{ijk}$ of the tensor corresponds to the entry related to the $i$-th user in the $j$-th feature at the $k$-th match in her/his gaming history. 

Once the dataset is represented in the tensor form, we can apply tensor decomposition techniques to perform a dimensionality reduction on the data. 
Here, we focus on the Non-negative Tensor Factorization (NTF) which is given by the PARAFAC/CANDECOMP decomposition with non-negative constraints~\cite{royer2011computing}.
The NTF approximates the tensor $\mathbfcal{X}$ into the sum of rank-one tensors, called components:
\begin{equation}
\mathbfcal{X} \approx \sum^{R}_{r=1} \lambda_r \mathbf{a}_r\circ\mathbf{b}_r\circ\mathbf{c}_r\;,
\label{ntf}
\end{equation}
where $R$ is the rank of the tensor, $\lambda_r$ are the values of the tensor core $\mathbfcal{L}= diag({\boldsymbol \lambda})$, and the outer product $\mathbf{a}_r\circ\mathbf{b}_r\circ\mathbf{c}_r$ identifies the component $r$, with $r=1,\dots,R$.

The vectors $\mathbf{a}_r, \textbf{b}_r$ and $\textbf{c}_r$ respectively provide the level of membership of the users to the component $r$, the level of membership of the features to the component $r$, and the temporal activation of the component $r$. These vectors can be encoded in three matrices $\mathbf{A}\in\mathbb{R}^{I\times R}, \mathbf{B}\in\mathbb{R}^{J\times R}$, and $\mathbf{C}\in\mathbb{R}^{K\times R}$, called factor matrices, whose $r$-th columns coincide to the vectors $\mathbf{a}_r, \textbf{b}_r$ and $\textbf{c}_r$. Therefore, the approximation in Eq.\eqref{ntf} can be rewritten in the Kruskal form~\cite{kolda2006multilinear} as
\[
\mathbfcal{X} \approx \llbracket\mathbfcal{L};\mathbf{A},\mathbf{B},\mathbf{C}\rrbracket\;.
\]
To obtain the factor matrices, and thus the approximated tensor, we need to solve an optimization problem of the form:
\begin{align*}
&\min \left\|\mathbfcal{X}-\llbracket \mathbfcal{L};\mathbf{A},\mathbf{B},\mathbf{C}\rrbracket\right\|^2_F \\
&\mbox{s.t.}\;\; {\boldsymbol \lambda}, \mathbf{A},\mathbf{B},\mathbf{C}\geq 0 
\end{align*} 
where $\|\cdot \|_F$ is the Frobenius norm, and a non-negativity constraints has been imposed on the factor matrices. There exist several algorithms to solve this optimization problem. Here, we rely on the Alternating Non-negative Least Squares (ANLS)~\cite{kim2007non}, combined with the Block Principal Pivoting (BPP) method, developed by~\cite{kim2012fast}.

\input{corcondia}

Once the number of components and the corresponding factor matrices are identified, we can analyze the information provided by the factor matrices. By the study of the matrices $\mathbf{A}$ and $\mathbf{B}$, we can respectively define the level of membership of each user to a specific component, as well as the level of membership of each feature to the components. It is worth noting that NTF allows users (or features) to belong to several components leading to overlapping groups, but also allows users (or features) to have a low level of membership such that it could be not enough to label users (or features) as members of a specific component. 

The are two possible strategies to define whether a user (analogously for features) belongs to a component: either we accumulate the membership of a component until the $95\%$ of its norm is reached~\cite{sapienza2015detecting}; or, we compute an intra-component k-means with $k = 2$ clusters~\cite{gauvin2014detecting}, i.e., for each component we divide the users in two groups: those who belong to the specific component, and those who do not. These methods allow to identify user clusters that might overlap. However, the use of such methods could lead to having users that do not belong to any component. 

In this work, we use the first of the two methods to analyze the level of membership of the features, provided by the matrix $\mathbf{B}$. This is done to detect the features having a key role in the different components. This decision is justified by our expectation that, in League of Legends (and in MOBA games in general), players can exhibit different strategies during each match, reflecting different personal goals: for example, a user may try to kill as many enemies as possible to earn a great amount of gold; this however would potentially incur in risking her/his hero's death more frequently; an alternative could be avoiding to get the hero killed too often by helping (assisting) other teammates, which however incurs in earning less gold at the end of the match.

However, since our aim is to identify groups of users with different behaviors, we decided to study the user membership as follows. We consider the components of $\mathbf{A}$ as observations recorded for each user, and we fit the k-means algorithm by imposing  $k = R$, i.e., we want to obtain one cluster of users for each component. In this way, the users are divided in $k$ disjoint clusters. 

Finally, we can recover the temporal activation of each component in the columns of the factor matrix $\mathbf{C}$. This information can be used to investigate the evolution of each extracted behavior. 

\section{Data collection and processing}

\subsection*{League of Legends}
To extract meaningful patterns related to how people behave in online platforms, and in particular to detect groups of users characterized by similar attitudes and characteristics, we tested the non-negative tensor factorization on a dataset of a MOBA game: League of Legends (LoL). League of legends is a match-based game developed by Riot Games, in which each player, i.e. user, controls a champion characterized by specific abilities and fight, together with other players, against a team of other players. The final goal is to defeat the adversary team in an arena. Each champion starts the match with a low strength level which increases by killing adversaries, helping members of the team in kills, i.e., assists, and performing other actions. During the match, each champion can be killed many times, i.e., number of deaths. The player can earn gold (i.e., the LoL currency) by performing some actions, such as killing or assisting in kills, and can use the earned gold to improve the abilities of the champion. 

We collected the LoL data by means of the Riot Games API,\footnote{\url{https://developer.riotgames.com/}} which provides metadata related to each match, including players' performance, such as number of assists, kills, deaths, etc. Each player is identified by a unique label and each match is marked by temporal information, such as match datetime and duration.

\subsection*{The League of Legends Dataset}
The dataset analyzed in the present work consists of $961$ players, and the complete game history of their first $100$ matches played exclusively in one specific same battle arena. We decided to focus on a specific battle arena to minimize the variability in players' behaviors (and their evolution) induced by different game scenarios. To this purpose, we selected the most popular LoL battle arena, namely the \textit{Summoner's Rift} (map\_id=11). This is (by a large margin) the most played battle arena in the game; its choice provided us with a significant amount of players who played this scenario at least 100 times. We decided to set this threshold because we wanted to guarantee that a sufficient number of matches were played by each single individual to capture a pattern of temporal behavior evolution. 100 matches resulted in a good trade-off between the number of users (nearly one thousand) and the number of total matches (nearly $100K$) yielded by the selected threshold.

\subsection*{Feature Selection}
For each match, the Riot Games API returns a very large number of features (50+) associated with the performance of each user involved in that match.\footnote{\url{https://developer.riotgames.com/api/methods}} 
Not all these features are predictive of player's game performance or behavior, and many of these features are highly correlated one another. To identify informative features, we designed a simple machine learning task: predicting whether a user had won a given match, given the vector of 50+ features describing her/his performance in that match. 

We use Decision Trees for this simple prediction task due to its ease of interpretation: feature can be ranked according to the information gain yielded by their addition to the model. 
The best model, which obtains a prediction accuracy above 80\%, selects the following four features as responsible for over 99\% of the information gain: \textit{(1)} number of assists, \textit{(2)} number of kills, \textit{(3)} number of deaths, and \textit{(4)} gold earned. We retain these four features and discard all others in the rest of the analysis. The final tensor is composed by $I=961$ users, $J=4$ features, and $K=100$ time steps.

The selected features are characterized by different ranges of values, thus the need to normalize them. Given the vector $\mathbf{x}_f$ related to the feature $f$, we normalize each entry $i$ of the vector as 
\[
\hat{x}_{f,i} = \frac{x_{f,i}-x_{min}}{x_{max}-x_{min}}\;,
\]
where $x_{min}$ and $x_{max}$ respectively are the minimum and maximum values of the vector $\mathbf{x}_f$.

\section{results}\label{sec:results}

Our League of Legends data is now represented as a tensor $\mathbfcal{X}^{I\times J\times K}$, where $I=961$, $J=4$, and $K=100$. Therefore, the resulting three-way tensor has dimensions related to the players, the selected features, and the time steps, which here coincide with the matches. Once the tensor is created, we compute its approximation $\mathbfcal{X}_{app}$, by applying  NTF.

To ensure of selecting the best approximation, we run $5$ simulations of the optimization problem with non-negative constraints and we repeat this procedure while varying the final number of components $r$, i.e., the rank. We performed the simulations for the rank values $r=1,\dots,10$ and selected the suitable rank on the basis of the results provided by the Core Consistency Diagnostic: the number of components that yields the largest knee in the slope of the core consistency curve is  $R=3$. We select this value of components for the following analysis.
By fixing the number of components, we then select the best approximation, by choosing the one corresponding to the maximum value of core consistency for the selected rank $R$.

\begin{figure}[t!]
  \includegraphics[width=0.5\textwidth]{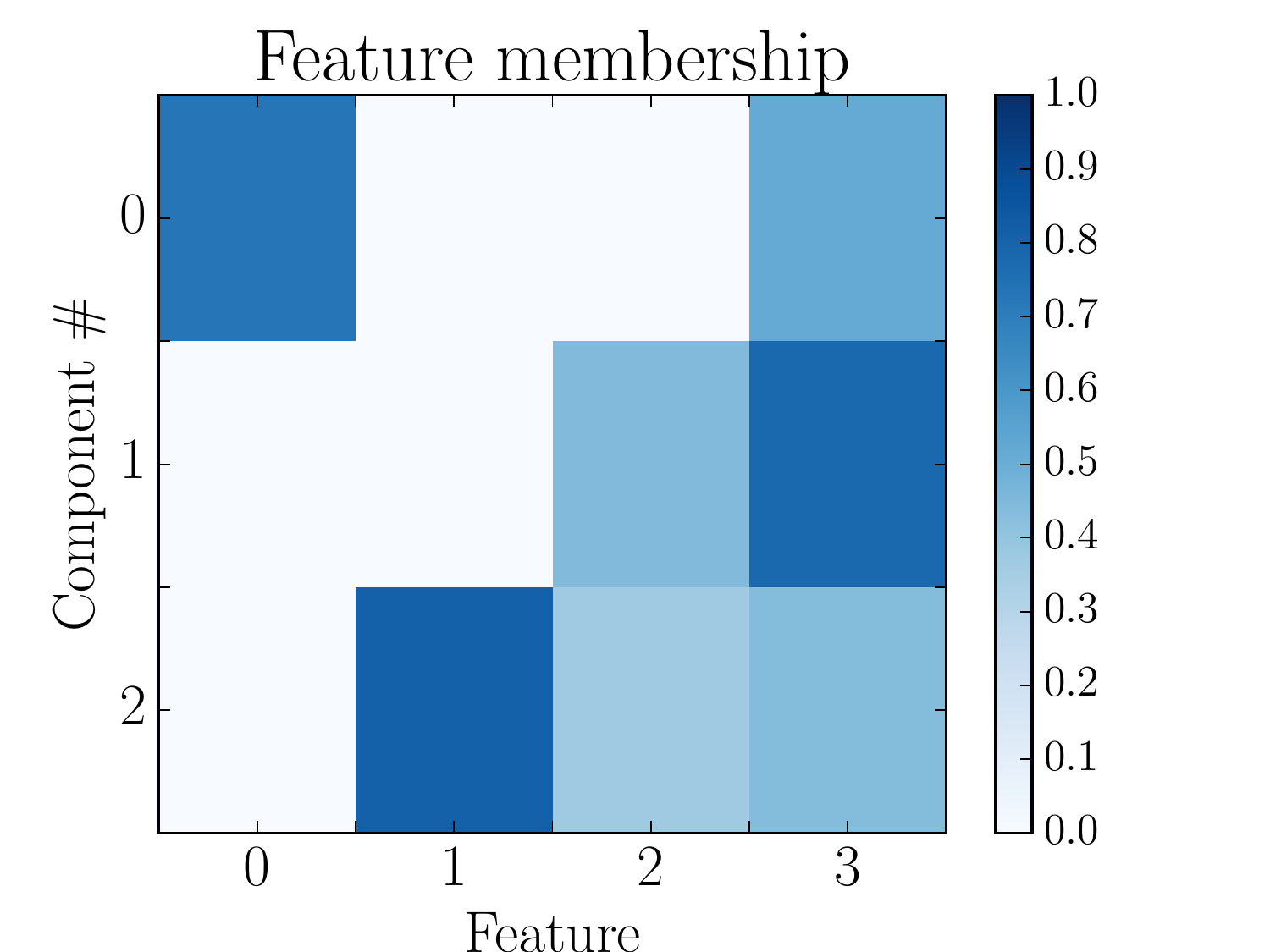}
  \caption{\textbf{Feature membership in the components}: we report the matrix $\mathbf{B}$ in which we zero-out the entries which are not included in the $95\%$ of the norm of the component. Here, the colorbar indicates the level of membership of each feature to the three different components. The features are respectively: assists ($0$), deaths ($1$), kills ($2$), and gold ($3$).}
\label{B}
\end{figure}

The approximated tensor is the output of the NTF and it is summarized in the factor matrices, as
\[
\mathbfcal{X}_{app} = \llbracket\mathbf{A},\mathbf{B},\mathbf{C}\rrbracket\;.
\]
We first analyze the results provided in the matrix $\mathbf{B}$, which are shown in Fig.~\ref{B}. Here, we report the values of the matrix $\mathbf{B}$ that have a key role in the components. To this aim, for each component we sort their squared values in descending order, sum them (starting from the highest value) until we reach the $95\%$ of the overall component norm, and set the remaining values equal to zero. The result of this procedure, shown in Fig.~\ref{B}, highlights the features that are involved in each component. Here, the features are marked as follows:

\begin{itemize}
\item \textbf{0:} number of assists;
\item \textbf{1:} number of deaths;
\item \textbf{2:} number of kills;
\item \textbf{3:} amount of earned gold.
\end{itemize}

The results provided by studying the factor $\mathbf{B}$ are reported in Tab.~\ref{tabfeat}, where we marked all the features that are involved in a certain component. As an example, the component $0$ is characterized by the features related to the assists and the earned gold.

\begin{table}[h!]
\centering
\caption{Features involved in each component.}
\label{tabfeat}
\begin{tabular}{c|c|c|c|c}
\textbf{} & Assists               & Deaths                & Kills & Gold           \\ \hline
Component 0                       & \checkmark &                       &       &  \checkmark                     \\ \hline
Component 1                       &                       &                       &    \checkmark   &                      \checkmark \\ \hline
\multicolumn{1}{l|}{Component 2}  & \multicolumn{1}{l|}{} & \multicolumn{1}{c|}{\checkmark} &   \checkmark    & \multicolumn{1}{c}{\checkmark}
\end{tabular}
\end{table}

Once the components are characterized through the features having the highest memberships, we analyze the results related to the users. The user memberships to each component are summarized in the factor matrix $\mathbf{A}$. As we have explained in Sec.~\S\ref{met}, we find the users belonging to each component by taking advantage of a k-means method with a number of clusters equal to the number of components, $k = 3$ in this case. This assumption is also supported by the Silhouette scores, computed for several values of $k$. In particular, for $k = 3$ the score is equal to $0.35$, while by increasing the number of clusters the score decreases, assuming values equal to $0.32$ and $0.29$ respectively for $k = 4$ and $k = 5$ and stabilizing below $0.29$ for $k > 5$. Thus, by fixing $k = 3$ the method assigns a unique label to each user and divides them into three disjoint groups.

\begin{figure*}[h!]
  \includegraphics[width=.92\columnwidth, clip=true,trim=0 10 50 25]{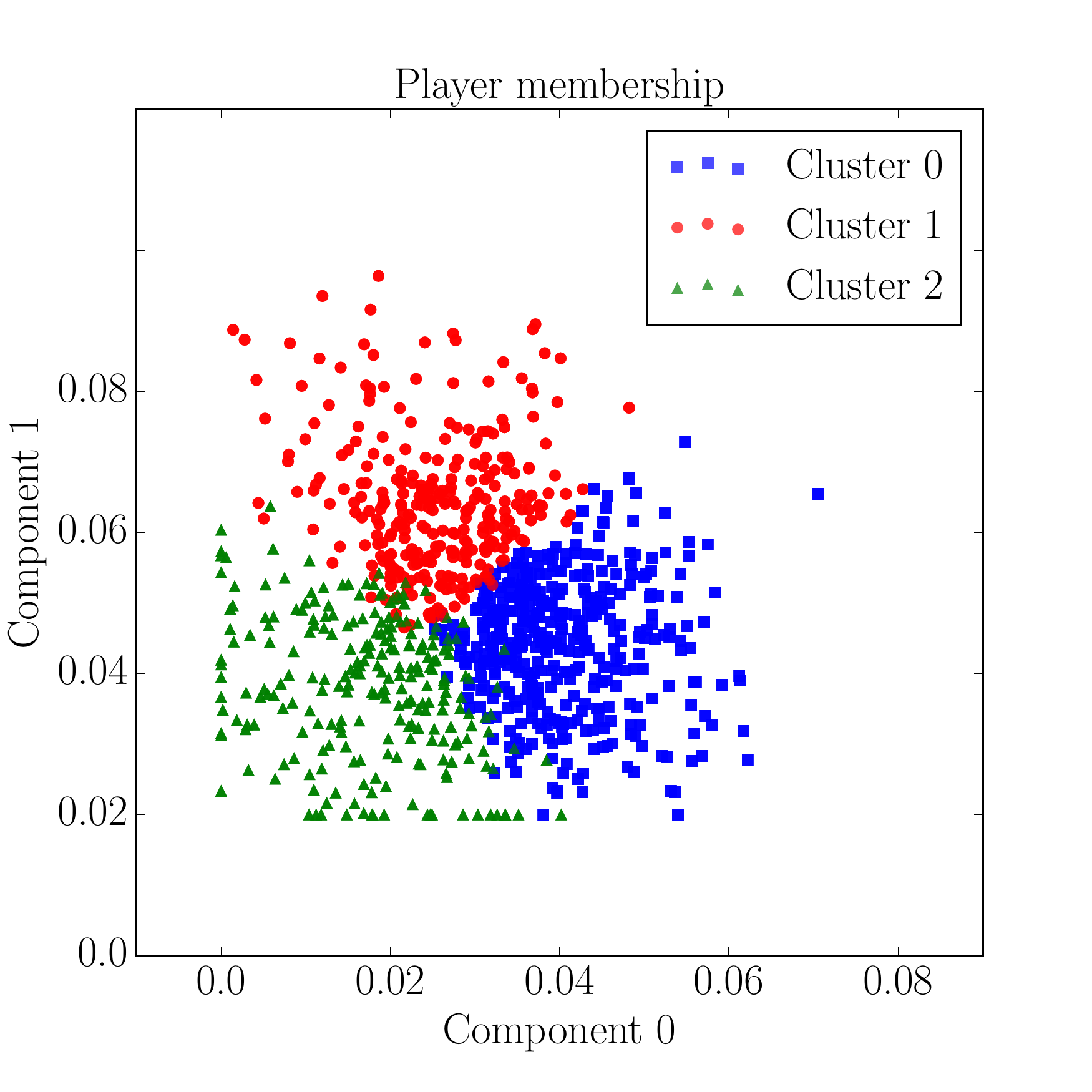}
  \includegraphics[width=1.17\columnwidth, clip=true, trim=0 0 0 0]{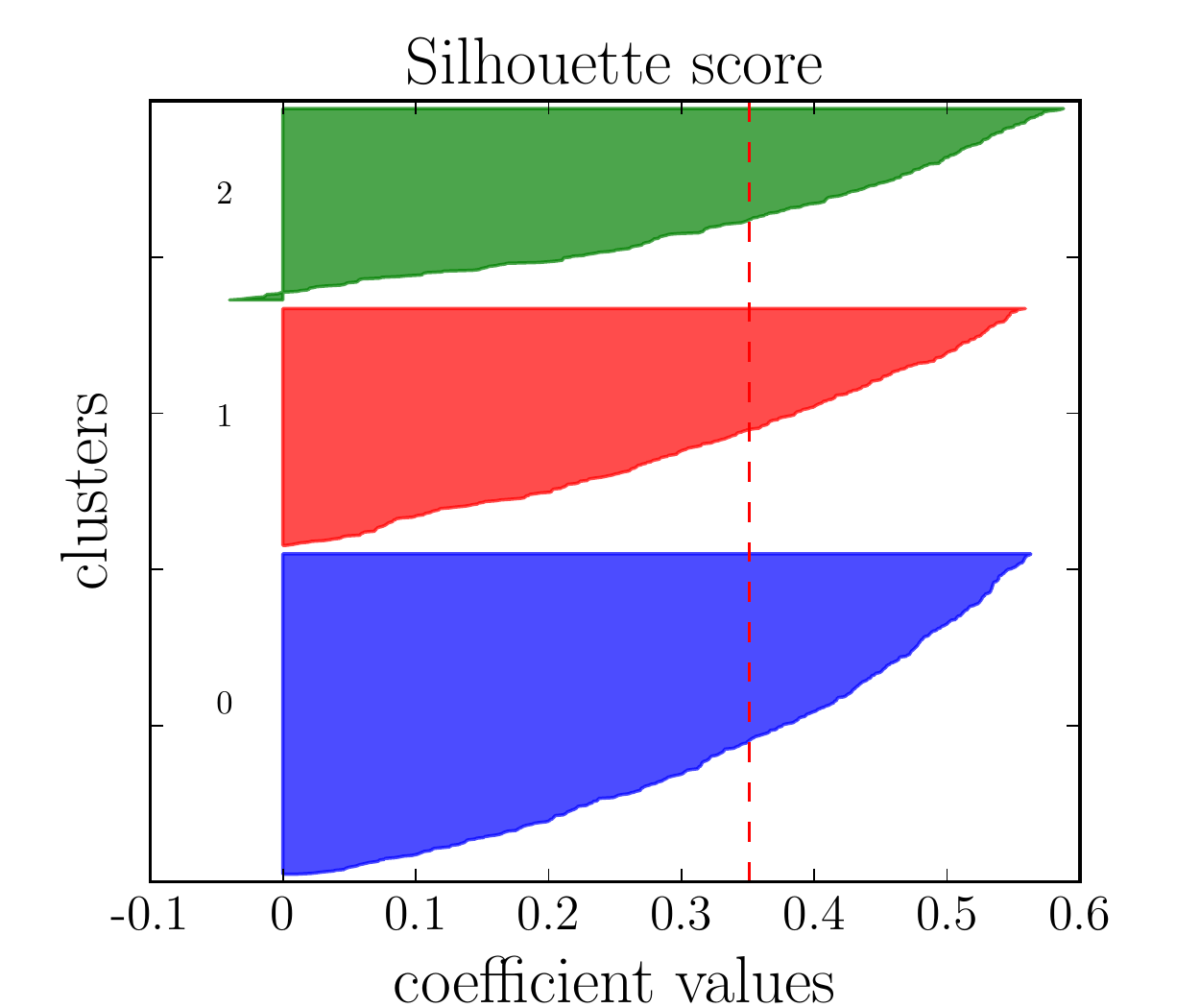}
  \caption{\textbf{K-means results:} the left figure shows the 2-dimensional projection of the three clusters identified by the k-means. Each dot represents a player in its corresponding cluster, and the dot's coordinates are given by the first two columns of the matrix $\mathbf{A}$. The right figure shows the Silhouette scores of the users belonging to the three clusters. The red line identifies the final Silhouette score of 0.35, and the width of the Silhouette profiles indicates the size of the corresponding clusters.
  }
\label{km}
\end{figure*}

As we can observe from Fig.~\ref{km}(left), the k-means with $k=3$ finds three clusters of users, which are disjoint in the component space, i.e., given a user $i$ its coordinates in the component space are given by the entries in the columns of the factor matrix $\mathbf{A}$. This grouping allows to identify the users whose membership to a specific component is higher than to the others. Clusters $0$, $1$, and $2$ respectively contain $411$, $304$, and $246$ users. Fig.~\ref{km}(right) shows the Silhouette profiles of the three clusters, proportional to their sizes.

\begin{figure}[h!]
	\begin{subfigure}{\columnwidth}
        \includegraphics[width=\columnwidth]{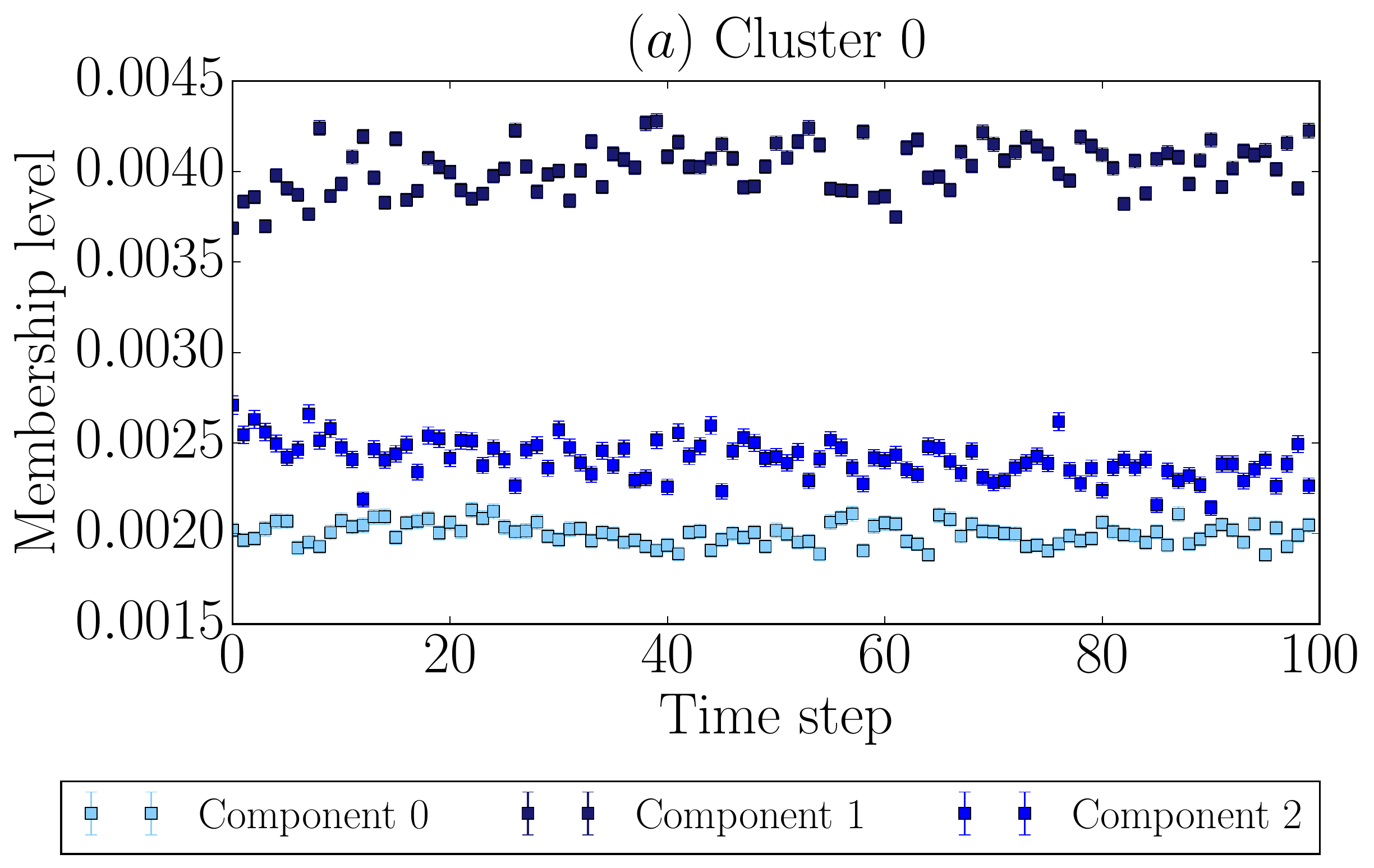}
        \label{ac0}
    \end{subfigure}
	\begin{subfigure}{\columnwidth}
        \includegraphics[width=\columnwidth]{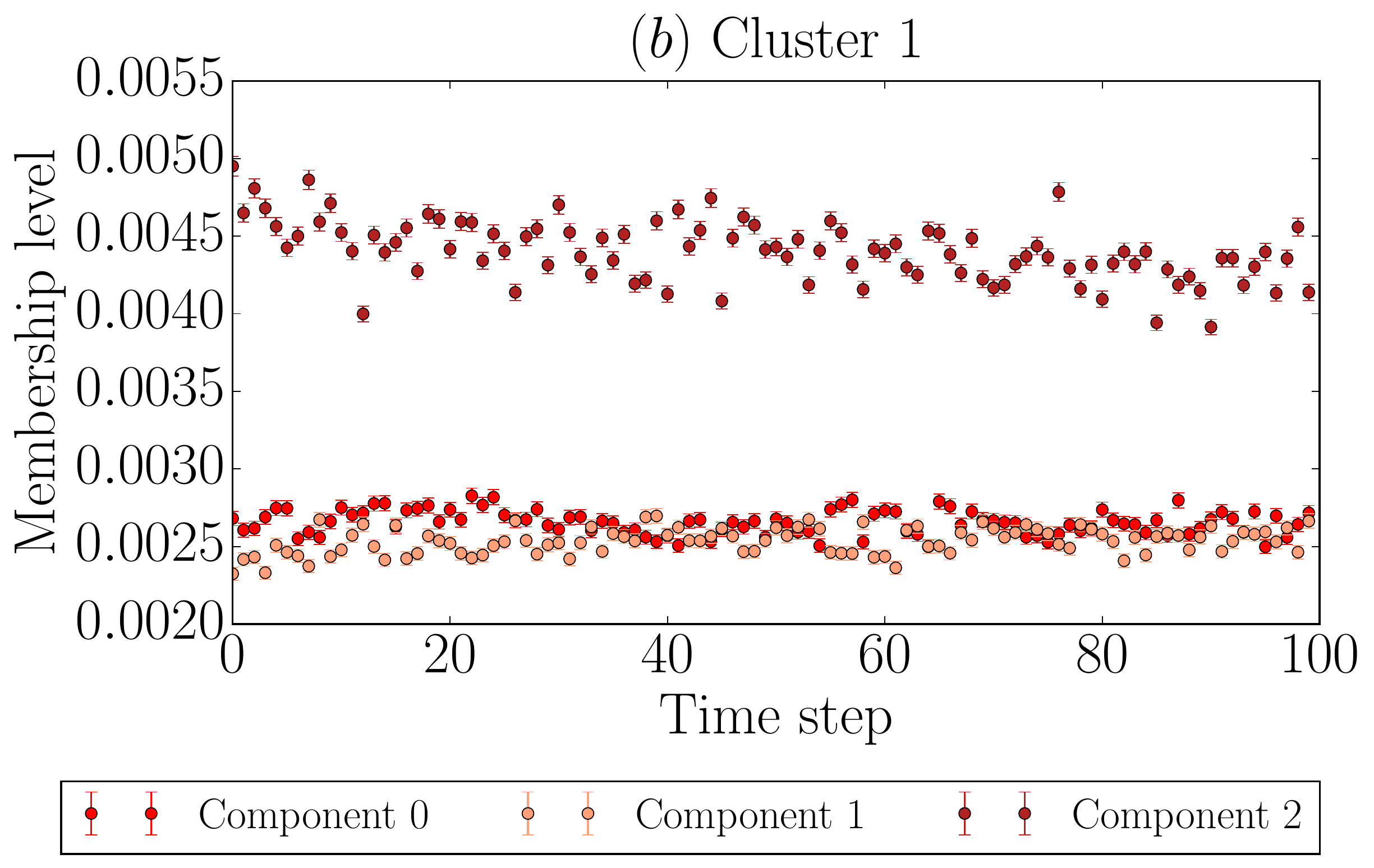}
        \label{ac1}
    \end{subfigure}
    \begin{subfigure}{\columnwidth}
        \includegraphics[width=\columnwidth]{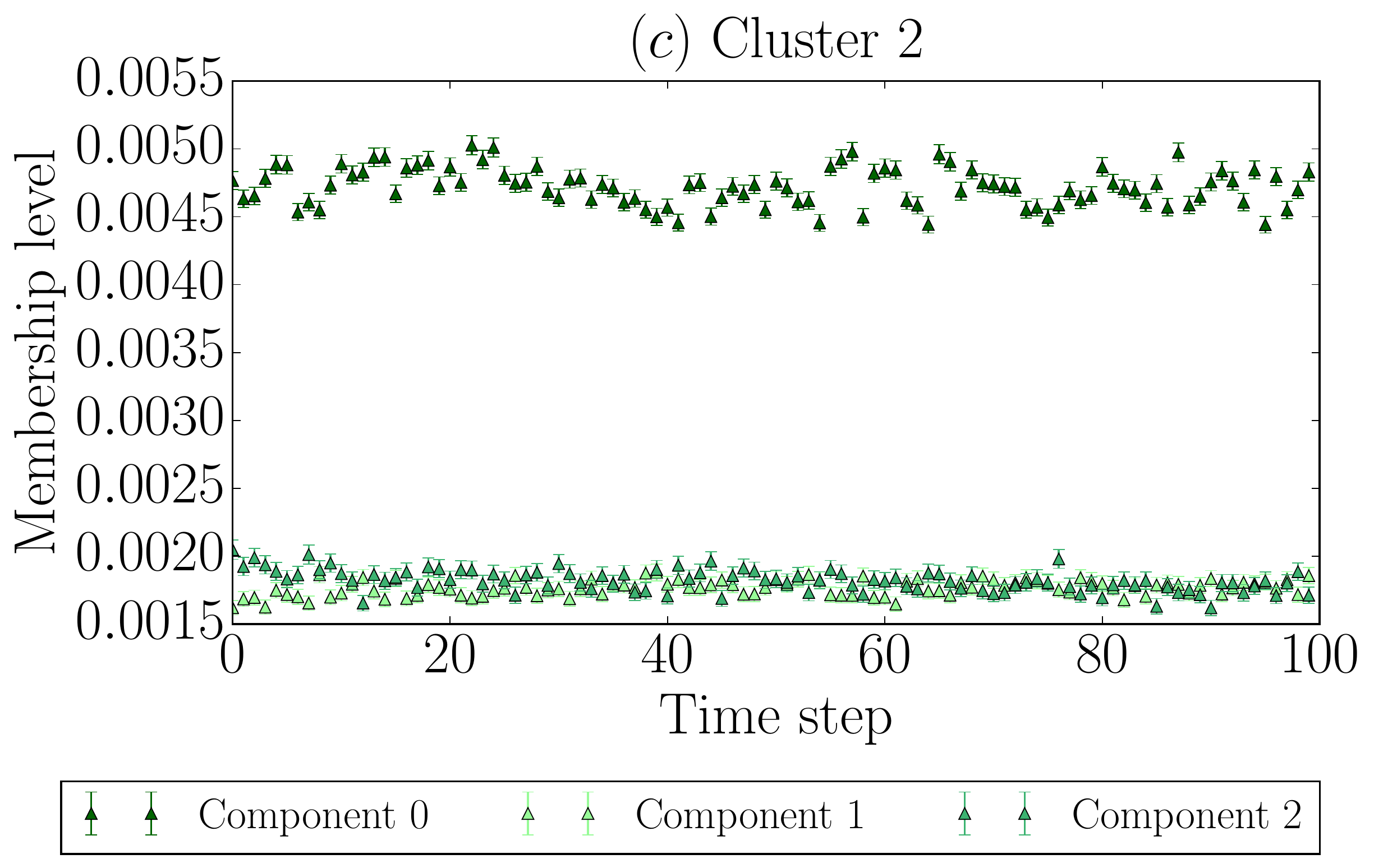}
        \label{ac2}
    \end{subfigure}
    \caption{Membership modulated in time. The figures display the product $\mathbf{a}_r\mathbf{c}^T_r$, which is the membership of users modulated in time. The product is computed by separately taking into account the users belonging to the different clusters. For each cluster we report the mean of the users membership to each component over time, and the related standard error, marked with an error bar. Different shades of blue for cluster $0$, red for cluster $1$, and green for cluster $2$ are assigned to distinguish the components.}
    \label{ac}
\end{figure}

The difference in the level of membership of users belonging to the different clusters is clear if we look at the values of $\mathbf{A}$ and how they are modulated in time. This is possible by computing for each $r$-th component the product $\mathbf{P} =\mathbf{a}_r\mathbf{c}^T_r\in\mathbb{R}^{I\times K}$, which represents the membership of each user to the $r$-th component modulated over time by the temporal activity of the $r$-th component. In Fig.~\ref{ac}, we report the average membership score over time and the related standard error (represented by error bars), computed by taking into account the users belonging to the same cluster in the three components. We can observe how each cluster is systematically characterized by an overall level of membership to one specific component that is much greater than to the other components. Fig.~\ref{ac}.a demonstrates the strong relation between Cluster $0$ and Component $1$ (\textit{cf.} black squares), Fig.~\ref{ac}.b shows the strong relation between Cluster $1$ and Component $2$ (\textit{cf.} dark red circles), while Fig.~\ref{ac}.c exhibits the strong relation between Cluster $2$ and Component $0$ (\textit{cf.}, dark green triangles). It is worth noting that there is an high gap between the average memberships over time to the cluster-related component and the remaining two components. This pattern indicates that:

\begin{itemize}
\item Users belonging to Cluster $0$ and thus to Component $1$ are strongly characterized by the features \emph{kills} and \emph{earned gold};
\item Users in Cluster $1$ and thus belonging to Component $2$ are characterized by \emph{deaths}, \emph{kills}, and \emph{earned gold};
\item Users belonging to Cluster $2$ and Component $0$ are strongly characterized by \emph{assists} and \emph{earned gold}.
\end{itemize}

Through the analysis of NTF results, we are able not only to identify the features that play a key role in a certain component, but we can easily find the user membership to the component. This enables to link each user in the component to the features that characterize the strategy used in the game. An interpretation for these results is indeed that different groups of users, jointly identified by NTF and k-means, are characterized by a playing behavior which is different from group to group. In particular, some users, such as those related to Component $0$, tend to collaborate more than others with their teammates, as they prefer to assists in fighting an enemy rather than killing him directly. Other users (e.g., Component $1$) are prone to perform individual actions, focusing on personal goals, such as earning a greater amount of gold, which can be spent to upgrade the player's champion abilities. Finally, in Component $2$ we detect a group of users that performs individual actions, such as a high number of kills, but are significantly more likely to cause their hero to die during these actions. This might pinpoint to a group of users characterized by an overall lower performance if compared with the other players.

\subsection*{Validation}
To validate the results obtained via NTF and the related interpretation, we selected the players in each cluster, and then we computed the mean and standard error of the different feature values at each time step (i.e., each match).

\begin{figure}[h!]
	\begin{subfigure}{\columnwidth}
        \includegraphics[width=.94\columnwidth,clip=true,trim=0 65 0 5]{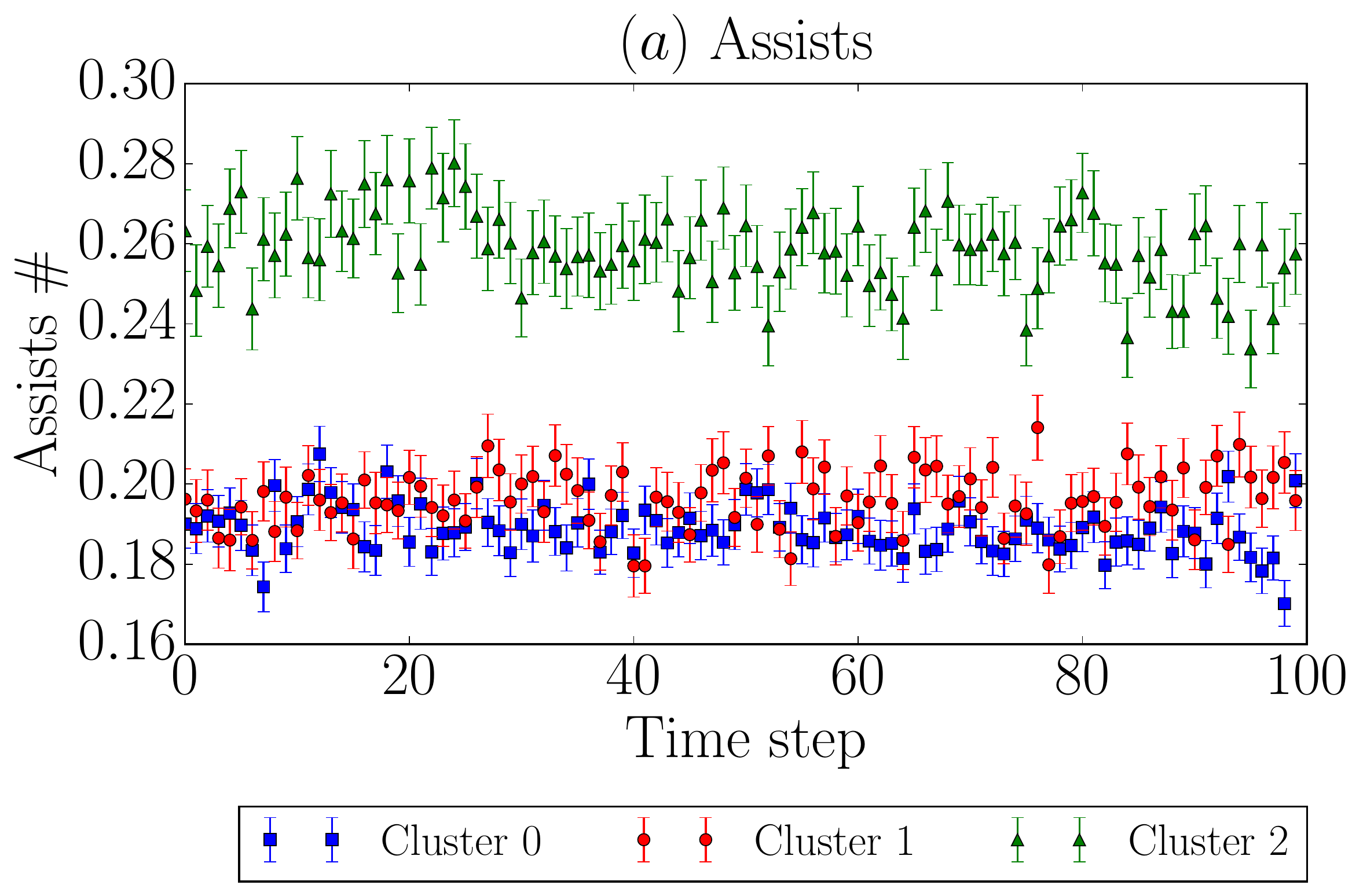}
        \label{ass}
    \end{subfigure}
	\begin{subfigure}{\columnwidth}
        \includegraphics[width=.94\columnwidth,clip=true,trim=0 65 0 5]{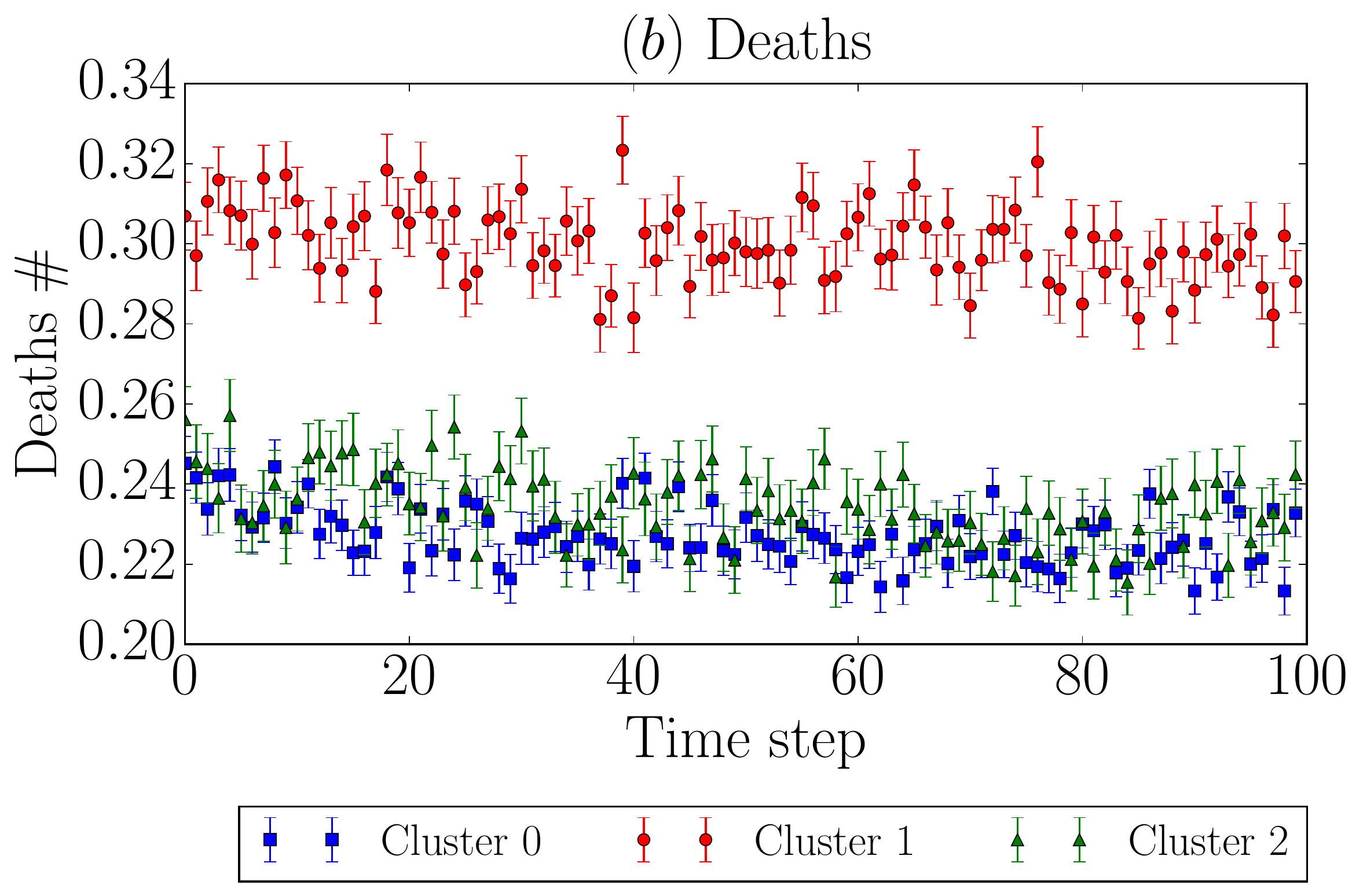}
        \label{dea}
    \end{subfigure}
    \begin{subfigure}{\columnwidth}
        \includegraphics[width=.94\columnwidth,clip=true,trim=0 65 0 5]{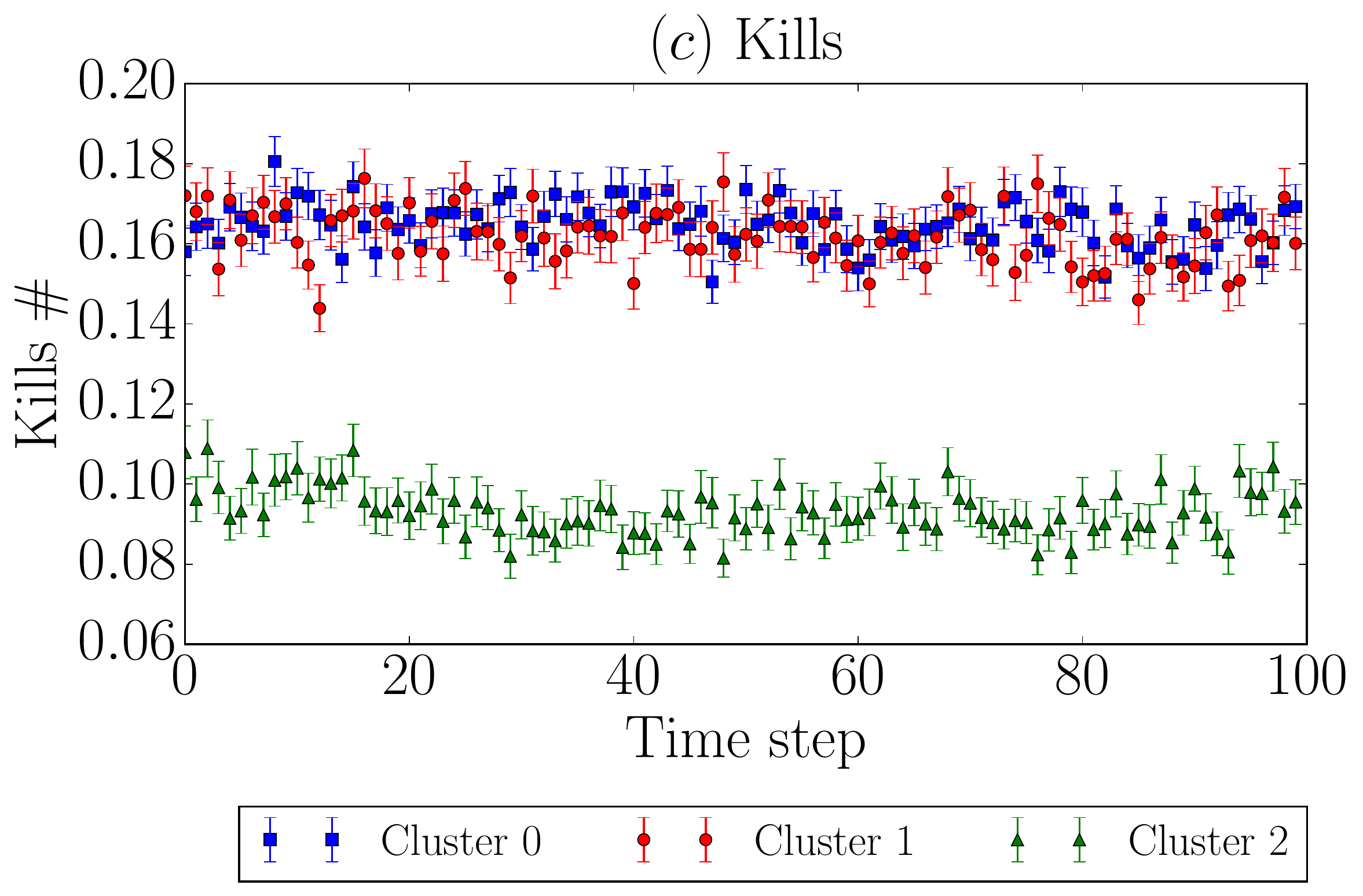}
        \label{kil}
    \end{subfigure}
    \begin{subfigure}{\columnwidth}
        \includegraphics[width=.94\columnwidth,clip=true,trim=0 0 0 5]{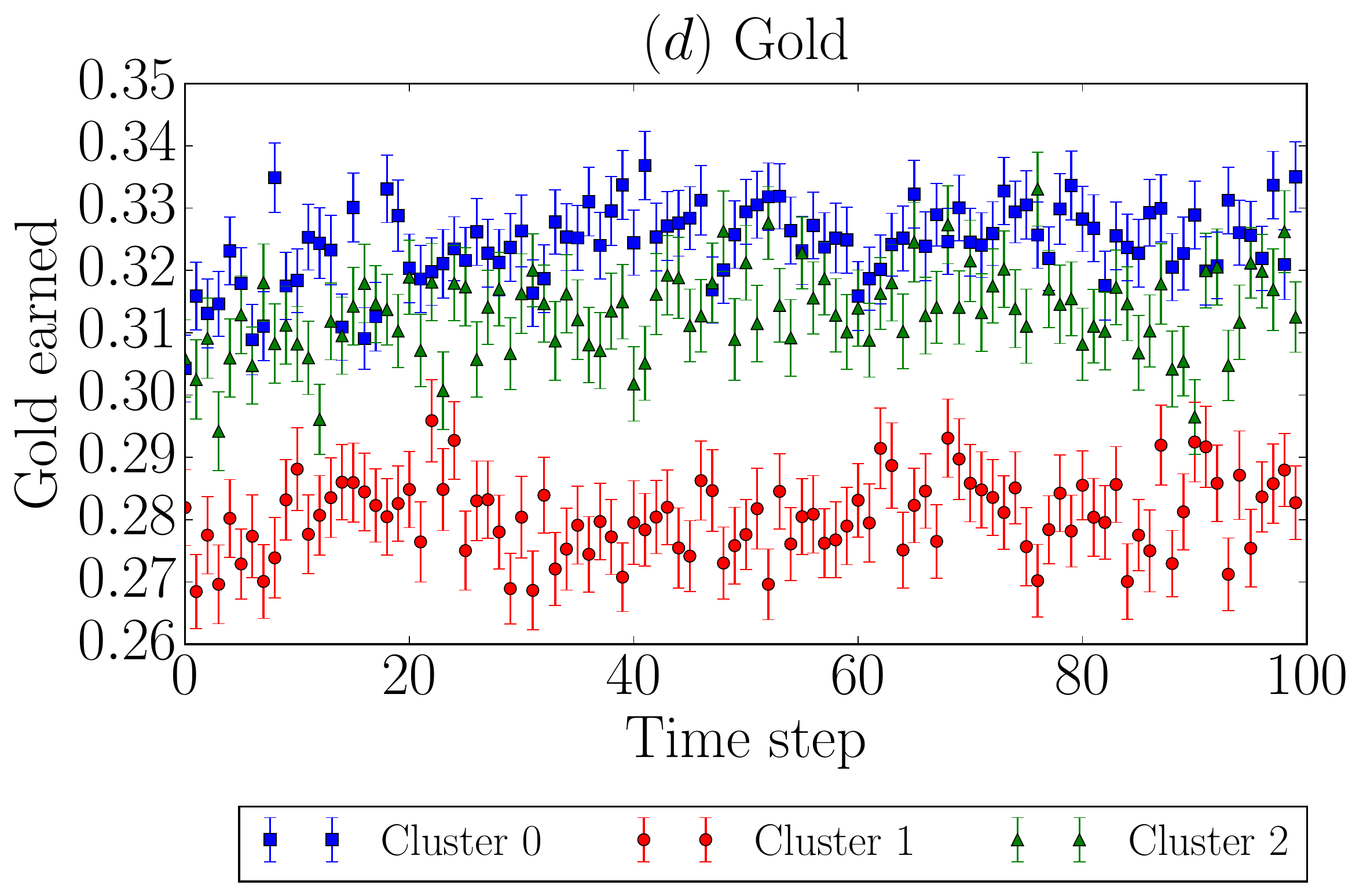}
        \label{gol}
    \end{subfigure}\vspace*{-.25cm}
    \caption{Mean feature values and related standard errors over time. The different subfigures show the player performance related to a certain feature over time. We computed the mean and the standard error over the values related to users belonging to the same cluster. Clusters are marked by a unique symbol and color, which is maintained in all the figures, to highlight the different cluster characteristics.}
    \label{feat}
\end{figure}

We show the results in Fig.~\ref{feat}, where each plot is related to a specific feature, namely \textit{(a)} number of assists, \textit{(b)} number of deaths, \textit{(c)} number of kills, and \textit{(d)} amount of earned gold. The results confirms the hypothesis and interpretation derived by the NTF analysis: Cluster $0$ (\textit{cf.,} blue squares) is composed by players whose major behavior dynamic over time is summarized by performing a high number of kills and earning at the same time a greater amount of gold than other players (as we mentioned earlier, the amount of gold is proportional to the number of kills, which here serves as a further sanity check). Cluster $1$ (\textit{cf.,} red circles) consists of players that obtain a number of kills comparable to the users in Cluster $0$, however their higher-than-average number of deaths causes them to systematically collect less gold (if compared with the other two clusters). Finally, Cluster $2$ involves players characterized by stronger social behavior, resulting in collaboration with the other team members, as conveyed by the larger number of assists and smaller number of kills. This strategy allows players in Cluster $2$ to collect a good amount of gold and at the same time to keep a low level of deaths.

One of the strengths of using a technique such as NTF is that we can disentangle the topological characteristics in the data, such as group of users characterized by similar features, from the temporal behaviors. The temporal information is indeed contained in the matrix $\mathbf{C}$, whose columns represent the timeseries of the temporal activation of each component, from which we can extract some meaningful interpretation of the evolution of players' behaviors. 

We expect that by testing different strategies, players can modify or adapt their way of playing to achieve better performances. This fact would be described by a change (such as an abrupt jump) in the temporal activity of a component, meaning that the component would activate or deactivate at a certain time.

However, by the analysis of the factor matrix $\mathbf{C}$, we can notice that each component is systematically active over time, i.e., despite different levels of activation, no significant behavioral change is noticeable in the behavioral trajectories of the players over the course of their 100 matches.

\begin{figure}[h!]
  \includegraphics[width=\columnwidth]{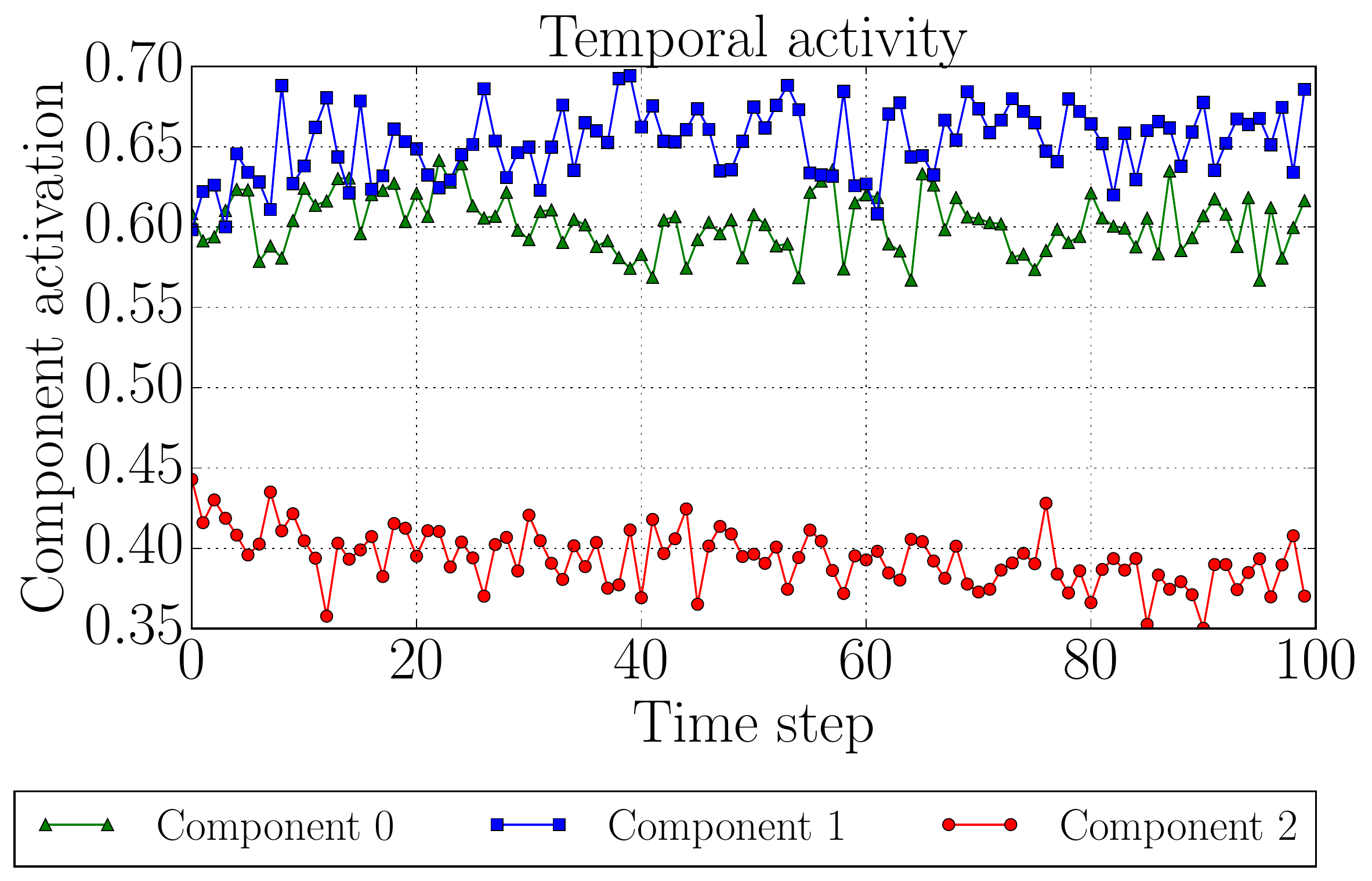}
  \caption{Temporal activity of each component. The values displayed in the figure are the one present in the columns of the factor matrix $\mathbf{C}$. The different markers characterize the different components, accordingly to the clusters colouring.}
\label{C}
\end{figure}

This result, illustrated in Fig.~\ref{C}, suggests that the group of users characterized by a specific strategy (i.e., one of the three leading strategies we highlighted above) is consistent over time; in other words, players are reluctant to continuously in changing their gaming behavior and strategy, even if that could occasionally entail a benefit.

We found an explanation for this phenomenon, which we suspect is related to the game design. League of Legends is based on a mathematical framework, that at the beginning of each match compares the players' skills to create the opposing teams as follows. Each player in LoL is characterized by an Elo-like rating\footnote{\url{https://en.wikipedia.org/wiki/Elo_rating_system}} which represents the player skill level, based on the performances in the previous matches. Thus, the resulting matchmaking rating is used by the system in assembling the teams and creating a game in which both teams have an equal chance of winning.\footnote{https://support.riotgames.com /hc/en-us/articles/201752954-Matchmaking-Guide}

\begin{figure}[h!]
  \includegraphics[width=\columnwidth]{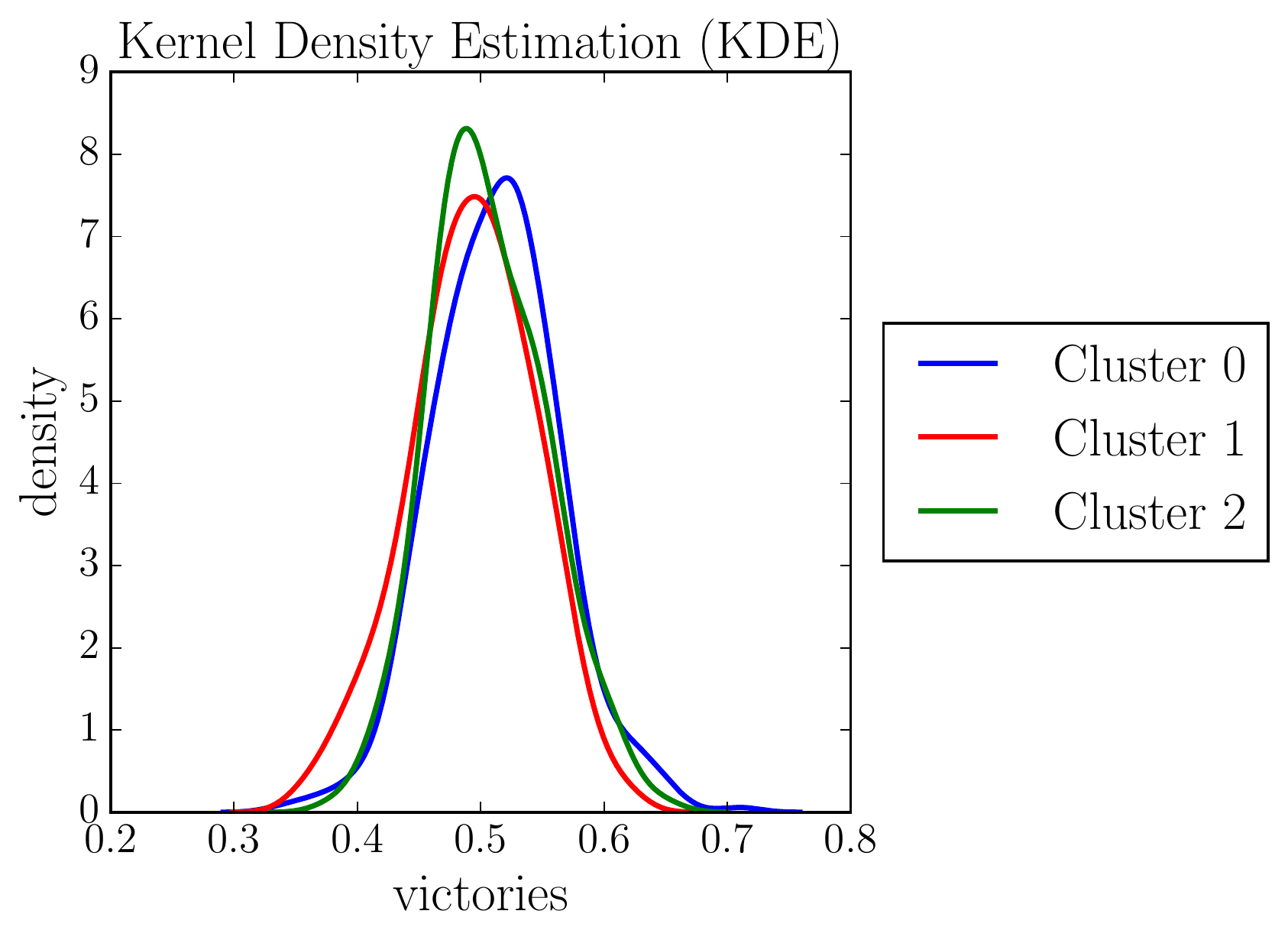}
  \caption{Kernel Density Estimation (KDE) computed on the values related to the feature \emph{winner} (binary feature equal to $1$ if a player wins the match and equal to $0$ if a player loses). The figure shows the probability density function for each cluster. We maintained the color code used throughout the text to discriminate the different clusters.}
\label{distr}
\end{figure}

Considering the LoL game design, we then compute the probability distribution of match winning for all players in each cluster identified by NTF. In Fig.~\ref{distr}, we report the Kernel Density Estimation (KDE) for the distribution of the victories in each cluster. The KDE estimates the probability density function of the feature \emph{winner} of each player involved in a certain cluster. This binary feature is equal to $1$ if a player wins a match, and $0$ otherwise.

By looking at the density function for each cluster, it appears that the distributions are centered around $0.5$, suggesting that each player at the beginning of each match has roughly a $50\%$ probability of winning or losing the match. However, closer inspection shows that these distributions are slightly skewed: Cluster $0$ leans toward values slightly higher than $0.5$, suggesting that the individualist strategy (aiming for a larger number of kills and less assists), on the long run yields marginally more victories than the cooperative strategy of Cluster $2$. We verified, using a pairwise two-tailed t-test, that these distributions are indeed statistically significantly different (all p-values  $\leq 10^{-3}$).

In conclusion, the League of Legends game design, and in particular the method used to create the opposing teams of a match, strongly affects the matches outcomes. Each user has roughly the same probability of winning a match, which is largely independent from the strategy used by players. Marginal changes can be noticed thanks to our NTF analysis that would otherwise get lost in the aggregated statistical analysis if oblivious of the social and temporal behavioral dynamics. We suggest that players have not incentives to change their natural behavior as they likely perceive to achieve the same winning performances of players characterized by a different strategy. This explains the temporal activity patterns discovered by NTF and their continuous and almost constant activation over time.

\input{related}

\section{conclusions}

League of Legends is a multiplayer online battle game in which two teams fight each other to destroy the respective enemy base. We collected data related to League of Legends matches and player performances with the aim of extracting meaningful information about human behavioral patterns. For this purpose, we took advantage of tensor decomposition techniques, a well established mathematical framework that is successfully applied to many research problems. In particular, we used the non-negative tensor factorization (NTF).

The advantage of using such a technique lies in the opportunity of disentangling the topological and temporal characteristics in the data, and exploring and validating them separately. 

Here, we analyzed a dataset composed by nearly one thousand players, characterized by different features, e.g. number of kills, number of deaths, etc., which varies over time, from match to match. We represented the data as a tensor and we applied NTF to extract the factor matrices related to the players, the features, and the temporal activities. 

The analysis of the NTF outcome and the application of clustering methods, such as the k-means, highlighted the presence in the data of several groups of players, characterized by a correlated behavior in time and topology. In particular, players belonging to the same component (cluster) are characterized by similar features and activation over time. 

We carried out the analysis of the topological characteristics of each group of player by looking at the features highlighted by NTF, and comparing the interpretation derived by these results with the original data. We found good agreement between the NTF output and the characteristics of the discovered groups of players in the original data. Therefore, NTF successfully identified groups of distinct behaviors in the data that can be interpreted as different player strategies. 

It would be expected that different strategies (e.g., collaborative vs. individualist playing) would lead to diverse performances (e.g., affecting the winning/losing ratio). However, by investigating the temporal activity patterns of the player groups, we found that they are mainly characterized by a constant behavior and that are active continuously over time. Thus, the analysis of the temporal activation of the NTF components stressed the reluctance of players to adapt their strategy and gaming behavior over time. 

This finding might be due to the game design of League of Legends:  the team formation in the game is based on a mathematical rule which aims at contrasting teams with comparable skills, thus yielding the same prior probability of victory to each team.

We confirmed this fact by computing the Kernel Density Estimation over the feature \emph{winner} for each player, divided by clusters. Only marginal, yet statistically significant, difference emerged, which are likely not perceivable by the players. Thus, players are not incentivized to changing their strategy with another one. 

In conclusion, the techniques and approaches used in this work are promising, and open new questions about human behaviors in multiplayer online games. Future work will be devoted to the analysis of additional game datasets with the aim of exploring behavioral patterns in different scenarios. The main goal would be to investigate if it is possible to nudge players to change their strategies by the use of incentives, such as rewards, and high percentage of victory.

\section*{Acknowledgments}
The authors are grateful to DARPA for support (grant \#D16AP00115). This project does not necessarily reflect the position/policy of the Government; no official endorsement should be inferred. Approved for public release; unlimited distribution.

\input{refs}

\end{document}

%% file: corcondia.tex
To find the best approximation, we run several simulations for each rank value. To assess the suitable number of components, i.e., rank, we use the Core Consistency Diagnostic (\textit{CORCONDIA}) by~\cite{bro2003new}: this method allows us to calculate the value of the core consistency for each simulation. In the present work, we run $5$ simulations for each rank and look at the change in the slope of the core consistency curve to select a suitable number of components $R$.

In particular, the core consistency values provide an evaluation of the closeness of the computed decomposition to the ideal one. This is done by comparing the core tensor $\mathbfcal{L}$ of the decomposition to a super-diagonal tensor $\mathbfcal{G}$, i.e., a tensor having entries equal to $1$ on the diagonal and $0$ otherwise. 

Therefore, the core consistency between $\mathbfcal{L}$ and $\mathbfcal{G}$ is defined as:
\[
cc = 100\left(1-\frac{\sum^R_{l=1}\sum^R_{m=1}\sum^R_{n=1}\left(\lambda_{lmn}-g_{lmn}\right)^2}{R}\right)\;,
\]
where $\lambda_{lmn}$ are the entries of the core tensor $\mathbfcal{L}\in\mathbb{R}^{R\times R\times R}$ and $g_{lmn}$ are the entries of the ideal core $\mathbfcal{G}\in\mathbb{R}^{R\times R\times R}$. 

The core consistency has an upper bound: its values cannot exceed $100$. As a result, high values of the core consistency mean high similarity between the cores, while low values indicate that the model selected could be problematic. Finally, the core consistency can assume negative values, thus indicating that the model selected is inappropriate. 
In the Results section (see \S\ref{sec:results}) we will discuss the optimal number of components as discovered by the the Core Consistency Diagnostic.

%% file: related.tex
\section{Related work}

Several facets of League of Legends, including players behaviors, expertise, and features have been already investigated~\cite{kou2014governance,donaldson2015mechanics}. Many works are focused on the analysis of League of Legends team composition. In~\cite{ong2015player}, the authors analyze player behaviors by developing a framework based on unsupervised learning techniques to discover behavior clusters in the data. In particular, they try to learn the optimal team composition and demonstrate how the result of matches can be predicted on the basis of the features characterizing the team. In our work, we used the winning prediction task to determine the most informative features that characterize players behaviors.

In~\cite{kou2014playing} the authors study the social interactions and organization patterns of LoL players, to understand how collaboration  arise during MOBA games. They collected a dataset based on interviews of experienced LoL players and found that team members collaborate and coordinate to reach the same goal and increase their performance in the game.

Other studies of LoL are focused on the analysis of specific game features, such as the usage of different characters (i.e., champions)~\cite{lee2015investigating}, or the player choice of a specific role with respect to the one selected by the other team members~\cite{kim2016proficiency}. The main goal in these studies is to investigate how roles and specific features have an impact on the players' performance, to recommend team design and evaluations that can be used by players when selecting a character. Our study, however, discovers in an unsupervised fashion the playing behaviors and the players' roles during the matches: we believe that our strategy could enrich the insights that game designers and analysts need to improve the game experience.

It is also worth noting that most of the existing studies are mainly focused on the analysis of LoL from a team-based perspective, to characterize groups performance. In the present work, we investigated individual player behaviors, and how that related to player performances at the level of single matches. In our analysis, we also highlighted the crucial importance of the temporal dimension. We monitored the evolution of features over time, to determine whether and how players learn or modify their strategy, and to detect if a common activity pattern can be found.

%% file: refs.tex
\balance
\bibliographystyle{ACM-Reference-Format}